\def\year{2019}\relax
\documentclass[letterpaper]{article} 
\usepackage{aaai19}  
\usepackage{times}  
\usepackage{helvet}  
\usepackage{courier}  
\usepackage{url}  
\usepackage{graphicx}  
\usepackage{amsmath,amssymb,amsfonts}
\usepackage{algorithmic}
\usepackage{graphicx}
\usepackage{textcomp}
\usepackage[symbol]{footmisc}

\usepackage{graphicx}
\usepackage{float}
\usepackage{pdfpages}
\usepackage{algorithm2e}
\usepackage{multirow}
\begin{document}
%
\title{Hardening Deep Neural Networks \textit{via} Adversarial Model Cascades
}

\author{
	Deepak Vijaykeerthy$^1$\thanks{These authors contributed equally to this work.} , 
	Anshuman Suri$^2$\footnotemark[1] , 
	Sameep Mehta$^3$, 
	Ponnurangam Kumaraguru $^4$
	\\ 
	$^{1,3}$ IBM Research \\
	$^{2,4}$ IIIT Delhi\\
	\\
	deepakvij@in.ibm.com,
	anshuman14021@iiitd.ac.in,
	sameepmehta@in.ibm.com,
	pk@iiitd.ac.in
}
\maketitle
\begin{abstract}
Deep neural networks (DNNs) are vulnerable to malicious inputs crafted by an adversary to produce erroneous outputs. Works on securing neural networks against adversarial examples achieve high empirical robustness on simple datasets such as MNIST. However, these techniques are inadequate when empirically tested on complex data sets such as CIFAR-10 and SVHN. Further, existing techniques are designed to target specific attacks and fail to generalize across attacks. We propose \textit{Adversarial Model Cascades (AMC)} as a way to tackle the above inadequacies. Our approach trains a cascade of models sequentially where each model is optimized to be robust towards a mixture of multiple attacks. Ultimately, it yields a single model which is secure against a wide range of attacks; namely FGSM, Elastic, Virtual Adversarial Perturbations and Madry. On an average, \textit{AMC} increases the model's empirical robustness against various attacks simultaneously, by a significant margin (of $6.225\%$ for MNIST, $5.075\%$ for SVHN and $2.65\%$ for CIFAR10). At the same time, the model's performance on non-adversarial inputs is comparable to the state-of-the-art models.

\end{abstract}

\section{Introduction} \label{introduction}
Deep neural networks (DNNs) have found widespread usage in areas such as computer vision, natural language processing, and computational biology. While their performance matches (or exceeds) human evaluations on the same tasks, sensitive applications such as autonomous navigation and computational finance require a deeper understanding of the predictions before their results can be considered trustworthy. However, since DNNs work by learning non-trivial representations of data, the intermediate representations and feature spaces of these networks have become increasingly complex. As a result, there is scope for vulnerabilities to be introduced into the networks, leading to a host of privacy and security concerns. For DNNs, one of the ways these vulnerabilities have been exploited is through adversarial examples. Adversarial examples, while being indistinguishable to humans from a clean example, can cause DNNs to produce incorrect and, in some cases, disastrous results.

Furthermore, adversarial examples that misguide one model are often successful in deceiving other models that are trained to perform the same task; irrespective of their architectures or the data sets used to train either of them~\cite{papernot2017practical}. Attackers may, therefore, conduct an attack with minimal to no knowledge about the target model, by training a substitute model to craft adversarial examples and nonetheless succeed in deceiving the target model.

Liu \textit{et al.} propose an ensemble-based approach to generate adversarial examples using an ensemble-based approach and claim high attack rates. They also show transferability of these targeted examples in a black-box setting, where the target model's weights and architectures are unknown~\cite{liu2016delving}. Bhagoji \textit{et al.}~\cite{bhagoji2017exploring} propose an alternate method to exploit DNNs via gradient estimation, which, by their results, performs nearly as good as white-box adversarial attacks, while outperforming the then existing state-of-the-art attacks on black-box models. Tramer \textit{et al.} use an ensemble of models to increase the robustness of the target model~\cite{tramer2017ensemble}. However, they use pre-trained models to transfer examples, and do not consider the scenario involving \textit{adaptive} adversaries. 

The above vulnerabilities have inspired a lot of research on securing neural networks against such attacks. Although a number of defence strategies have been proposed in the recent past, many of them can be broken when the adversary adapts the attack to consider the defence. The most prominent among these strategies is \textit{adversarial training}~\cite{madry2017pgm} where the model is trained with adversarial samples generated by attacks, along with the standard set of training examples.  One of the crucial shortcomings of this approach is that due to over-fitting on samples from stronger attacks, the model may still be vulnerable to adversarial examples from weaker attacks~\cite{goodfellow2018}. In turn, this hinders the above methods from generalizing to multiple attacks.

It is also worth mentioning that there have been attempts to construct models which are secure by construction. This is achieved by exploiting the structural properties of DNNs. Owing to additional operations for such structural changes, these approaches are computationally expensive and scale only to models on simpler data sets such as MNIST.~\cite{aditi2018}, ~\cite{wong2018}, ~\cite{aman2018}.
Although these approaches increase the robustness of a model for simpler data sets, these strategies are ineffective for complex datasets such as CIFAR-10 and SVHN.

In this work, we build on the \textit{adversarial training} framework, and design a training technique called \textit{Adversarial Model Cascades (AMC)}. One of the key contributions of our paper is that, in addition to the traditional white-box setting, we also consider an \textit{adaptive} black-box adversary that makes use of queries to the target model's prediction function to train a proxy model. We demonstrate that \textit{AMC} improves the prior art’s empirical worst-case accuracy for several data sets namely MNIST, CIFAR10 and SVHN.

In addition to the above, we also investigate the suitability of Feature Squeezing~\cite{xu2017feature} as a pre-processing technique in the pipeline to improving the empirical robustness of DNNs. It is important to note that Feature Squeezing as a stand-alone defence has been shown to be easily bypassed~\cite{yash2018}. In contrast, when used in conjunction with our framework, we observe that Feature Squeezing effectively improves the accuracy against stronger attacks that the model has seen during training. Our work is the first to demonstrate an end to end pipeline to improve the empirical robustness of DNNs against multiple attacks simultaneously.


\section{Background}




\subsection{Threat Model}
Consider a DNN $f(.; \theta)$ and a clean (no adversarial noise) sample ($\vec{x}$, $y$) $\sim \mathcal{D'}$, where $\mathcal{D'}$ is a proxy for an unknown distribution $\mathcal{D}$, from which the set of samples used to train the target model $f(\vec{x}; \theta)$ are drawn. An adversary tries to create a malicious sample $\vec{x}_{adv}$ by adding a small perturbation to $\vec{x}$, such that $\vec{x}$ and $\vec{x}_{adv}$ are close according to some distance metric (either $L_1$, $L_2$ or $L_{\infty}$ norm), and $f(\vec{x}_{adv}) \neq y$. In each domain, the distance metric that we must use is different. 
In the space of images, which we focus on in this paper, we rely on previous work that suggests that various $L_p$ norms are reasonable approximations of human perceptual distance~\cite{carlini2017l2attack}.

For our analyses, we consider two threat-model settings: white-box and adaptive black-box. A white-box adversary is one that has access to the target model's weights and the training data set. An \textit{adaptive} black-box adversary is one that interacts with the target model only through its predictive interface. This \textit{adaptive black-box} adversary trains a proxy model $f_{P}(\vec{x}'; \theta')$ on the set $\mathcal{S}' = \{(\vec{x}'_i, y'_i)\}_{i = 1}^{n}$ , where $\vec{x}'_i$'s are drawn i.i.d from distribution $\mathcal{D}'$ and $y_i = f(\vec{x}'_i; \theta)$. In this setting, the adversary crafts adversarial examples ${\vec{x}}_{adv}$ on the proxy model $f_{P}(\vec{x}'; \theta')$ using white-box attack strategies and uses these malicious examples to try and fool the target model $f(\vec{x}; \theta)$. 





\subsection{Adversarial Examples} \label{attacks}
Given a clean sample $\vec{x}$ it is often possible to find a malicious sample $\vec{x}_{adv}$, such that $f(\vec{x}_{adv}; \theta) \neq y$, and $\vec{x}_{adv}$ is close to $\vec{x}$ according to some distance metric (either $L_1$, $L_2$ or $L_{\infty}$ norm). 

For our analyses, we consider the following attacks:
\begin{itemize}
	\item \textbf{Fast Gradient Sign Method (FGSM)}: Uses the gradient of a modified loss function to modify samples, constrained by $L_{\infty}$ norm~\cite{goodfellow2014explaining}.
	\item \textbf{Virtual Adversarial Perturbations (VAP)}: VAP perturbs $\vec{x}$ in the direction that can most severely damage the probability that the model correctly assigns the label $y$ (the correct label) to $\vec{x}$~\cite{miyato2016vae}.
	\item \textbf{Elastic Adversarial Perturbations (EAP)}: EAP is based on elastic-net regularization, which uses a mixture of $L_1$ and $L_2$ penalty functions~\cite{chen2017eadattack}.
	\item \textbf{Projected Gradient Method (PGM)}: Uses projected gradient descent (PGD) on the negative loss function~\cite{madry2017pgm}.
\end{itemize}

\begin{figure}[h]
	\centering
	\includegraphics[width=0.4\textwidth]{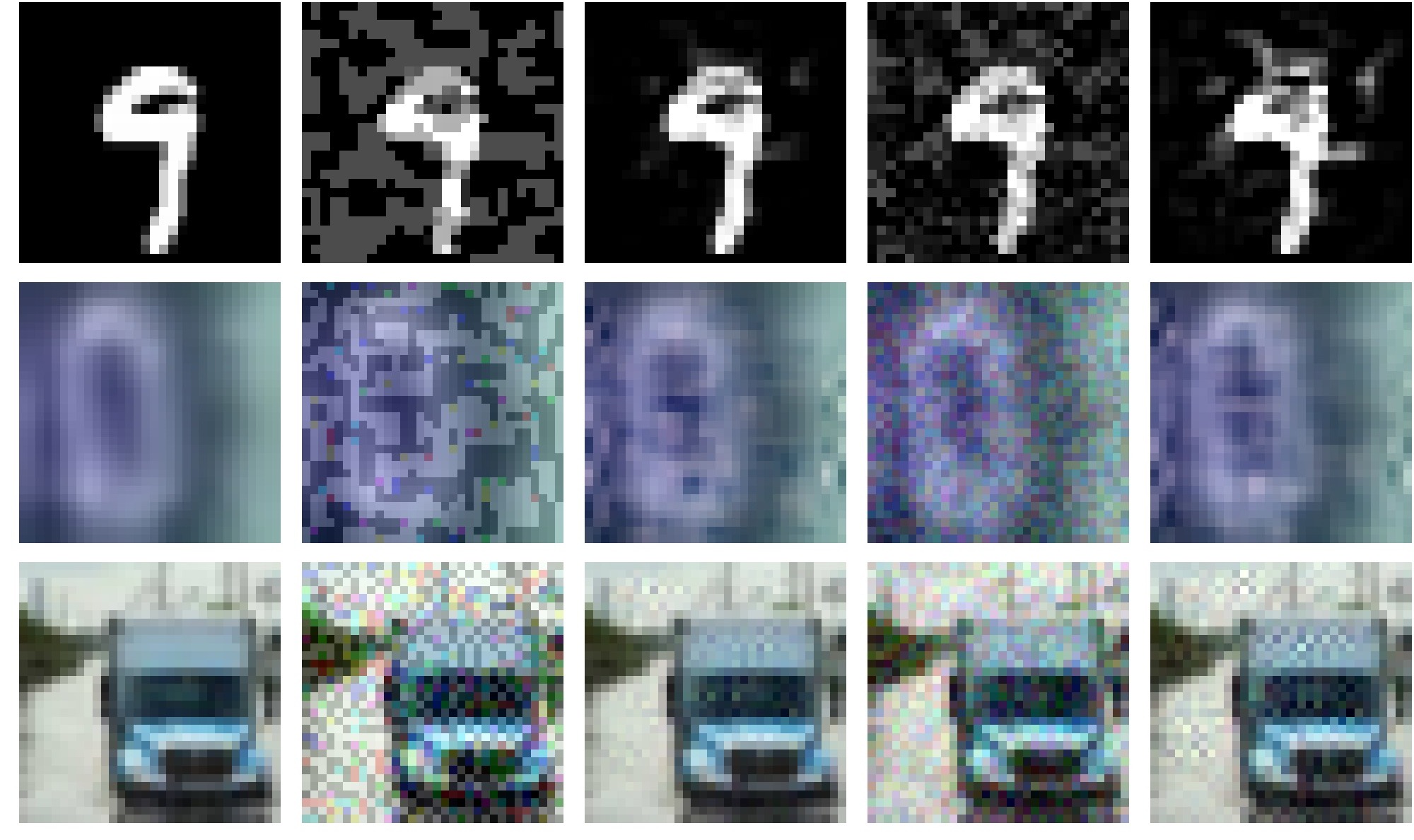}
	\caption{Examples with various perturbations generated via various attacks for all three datasets. All of these perturbed images make the target model misclassify. L to R: Unperturbed, FGSM, EAP, PGM, VAP. Top to Bottom: MNIST, SVHN, CIFAR-10.}
	\label{fig:attack_images}
\end{figure}

These four attacks are picked so as to cover the spectrum of adversarial attacks in literature. Examples for each of these attacks are shown in Figure \ref{fig:attack_images}.
One can also characterize attacks based on their strength, i.e. how easy or hard it is to defend against adversarial examples generated by the attacks. For example, VAP and FGSM are generally regarded as weak attacks as many defences proposed in literature have been demonstrated to be effective against VAP and FGSM attacks, as opposed to PGM which is harder to defend against.

Mathematically, all attacks can be viewed as solving the same optimization problem: an adversary tries to create a malicious sample $\vec{x}_{adv}$ by adding a small perturbation to $\vec{x}$, such that $\vec{x}$ and $\vec{x}_{adv}$ are close according to some distance metric (either $L_1$, $L_2$ or $L_{\infty}$ norm), and $f(\vec{x}_{adv}; \theta) \neq y$. Thus, the strength of an attack can be measured by how closely it can approximate the true optima of the above optimization problem.

\subsection{Defences}
Although many defence strategies to secure DNNs against adversaries have been proposed in the recent past, many of them can be broken when the adversary adapts the attack to take the defence into consideration. The most prominent among these strategies is \textit{adversarial training}~\cite{shaham2015adversarialtraining} where the model is trained with adversarial samples generated by attacks along with the standard set of training examples.
 
\subsubsection{Adversarial Training}
Adversarial training hardens the model against malicious examples by either re-training the model on an augmented set containing the training data and the adversarial examples or learning using the modified objective function:
\[
\hat{J}(\theta; \vec{x}, y) = \alpha.J(\theta; \vec{x}, y) + (1 - \alpha).J(\theta; \vec{x}_{adv}, y)
\]
where $J(\theta; \vec{x}, y)$ is the original loss function and $\alpha$ is a tunable hyperparameter. This defence aims to increase the model's robustness by ensuring that it predicts the same class for a clean example and its corresponding example with adversarial perturbation. 

As opposed to the general practice of adding adversarial examples without replacement, we add them with replacement. This helps us achieve better generalization performance over an unseen test-set.

\begin{algorithm}
\SetAlgoLined
\SetKwInOut{Input}{input}
\Input{An attack $\mathcal{A}$, training set $\mathcal{D}$, total number of epochs N, a model with parameters $\theta$, and learning rate $\eta$}
\Begin{
 \For{$i \leftarrow 1$ \KwTo N}{
   		Generate adversarial examples for samples in $\mathcal{D}$ using $\mathcal{A}$, $\{\vec{x}^{adv}\}$ \\
        $\theta \leftarrow \theta -  \nabla_{\theta}\hat{J}(\theta; {\vec{x}\cup \vec{x}^{adv}}, y)$
 }
 }
 \caption{A high-level overview of the \textit{the adversarial training} procedure \texttt{AT}($\mathcal{A}$, $\mathcal{D}$, N, $\eta$, $\theta$)} \label{alg:at_algo}
\end{algorithm} 

Intuitively, the algorithm iteratively does the following two steps
 \begin{itemize}
     \item Find the optimal $\vec{x}_{adv}$ such that $f(\vec{x}_{adv}; \theta) \neq y$
     \item Optimizes $\theta$, over the worst case adversarial example $f(\vec{x}\cup \vec{x}_{adv}^{*}; \theta) \neq y$
 \end{itemize}
 
Many of the earlier formulations of \textit{adversarial training} have been easily bypassed in practice; this has been attributed to the sharpness of the loss around the training examples. Only until recently, Madry \textit{et. al.} ~\cite{madry2017pgm} showed that adversarial training could be used to obtain a robust MNIST model. That is, the worst-case accuracy on MNIST adversarial examples could be no lower than 81\%, even though the attacker knows all parameters of the model. Although their results are encouraging, their technique still falls short in three aspects:
\begin{enumerate}
    \item it is ineffective against \textit{adaptive} black-box adversaries~\cite{tramer2017ensemble}, ~\cite{goodfellow2018},
    \item it is ineffective for complex datasets such as CIFAR-10 and SVHN.~\cite{madry2017pgm}, and
    \item it fails to generalize across attacks~\cite{goodfellow2018}.
\end{enumerate}

\section{Approach}

To address this challenge, we propose \textit{Adversarial Model Cascades (AMC)}  which can help secure models against \textit{adaptive} black-box attacks. Our approach is to train a cascade of models by injecting images crafted from an already defended proxy model (or from the target model itself) to improve the robustness against adversarial attacks. In the next section, we describe the critical steps involved in our approach; namely the construction of the proxy models and the adversarial model cascades. 

\subsection{Proxy Models} \label{prxy_models}
One of the critical steps in our proposed approach is to train a proxy (or surrogate) model to mimic the target model. The strategy is to train a proxy for the target model using unlabeled examples from the proxy distribution $\mathcal{D}'$. The labels for these data points are obtained by observing the target model\textquotesingle s output on these examples. Then, adversarial examples are crafted for this proxy. We expect the target model to misclassify them due to transferability between architectures~\cite{papernot2017practical}.

One may believe that the choice of a neural network architecture plays a vital role in the effectiveness of the proxy model and the adversary might find it hard to decide on a suitable one. However, the adversary has some partial knowledge of the oracle input (e.g., images, text) and expected output (e.g., classification) at the very least. The adversary can thus use an architecture adapted to the input-output relation. Adversaries can also consider performing an architecture exploration and train several substitute models before selecting the one yielding the highest misclassification. In our research, we use an architecture similar to one of those proposed by Urban \textit{et al.}, which they have shown to be effective in training surrogate models (via distillation) and replicating the predictive performance of the target model~\cite{ubran2016extraction}.

\subsection{Adversarial Model Cascades} \label{amc}
 Inspired by the observation that adversarial examples transfer between defended models, we propose \textit{Adversarial Model Cascades(AMC)} which trains a cascade of models by injecting examples crafted from a local proxy model (or the target model itself). The cascade trains a stack of models built sequentially, where each model in the cascade is more robust than the one before. The key principle of our approach is that each model in the cascade is optimized to be secure against a combination of the attacks under consideration, along with the attacks the model has encountered during the previous iteration. Knowledge from the previous model is leveraged via parameter transfer while securing the model against subsequent attacks. This technique increases the robustness of the next layer of the cascade, which ultimately yields a model which is robust to all the attacks it has been hardened against via the algorithm. A high-level overview of the \textit{AMC} framework is summarized in Algorithm \ref{alg:amc_algo}.
 
\begin{algorithm}
\SetAlgoLined
\SetKwInOut{Input}{input}
\Input{Undefended model $\textrm{M}^0$, the training set $\mathcal{D}$ of size $N$ on which $\textrm{M}^0$ was trained on, a set of attacks $\mathcal{S}$ to harden against and the number of attacks E}
\Begin{
 \For{$i \leftarrow 1$ \KwTo E}{
   		Initialize model parameters of $\textrm{M}^{i + 1}$ to $\theta^{\textrm{M}^{i}}$\ \\
   		$\theta^{M^{i + 1}} \leftarrow$ \texttt{AT}($\mathcal{S}(i)$, $\mathcal{D}$, N, $\eta$, $\theta$)
 }
 Predict using the final model in the cascade : $\hat{y} = \arg\max_{y}\textrm{M}^{E}(x)$\;
 }
 \caption{A high-level overview of the \textit{AMC} framework. The algorithm works by transferring knowledge for a specific attack and building upon it iteratively to increase robustness.} \label{alg:amc_algo}
\end{algorithm} 

In the case of a white-box adversary, adversarial examples $X_{adv}$ are crafted for the corresponding model $\textrm{M}^{i}$ (\textit{AMC, target-hardened}). For the case of an \textit{adaptive} black-box adversary (\textit{AMC, local proxy}), adversarial examples are crafted using an identical local proxy model $\textrm{P}'_{i}$: the proxy adversarially perturbs training data. This data is then concatenated with normal training data by the target model for training. It is important to note that, for our approach to work in an \textit{adaptive} black-box scenario, we need access only to the prediction interface of model $\textrm{M}^{i}$.

\subsubsection{Forgetting Attacks}
Although \textit{AMC} achieves a significant increase in robustness against many attacks, it does not guarantee an increase in robustness against all of the attacks which it has seen in the past. This condition is especially true for attacks which were seen during the initial iterations of the algorithm. To mitigate this problem, during the later iterations we also need to make the model remember the adversarial examples generated by attacks which were seen by it during its initial iterations. Thus, while constructing adversarial data per batch: instead of generating all perturbed data using the current attack, the algorithm also uses attacks it has seen so far into its run. This process implicitly weighs the attacks, as the model ends up seeing more samples from the attacks which it encounters during earlier iterations. As a result, the loss function to compute its gradient (at a given level $E$ while running the cascade) becomes:
\[
\hat{J}(\theta; \vec{x}, y) = \alpha J(\theta; \vec{x}, y) + (1 - \alpha) \sum_{i = 0}^{E} \lambda_{i} J(\theta;  \vec{x}^{i}_{adv}, y)
\]
where $\lambda_i$ are hyper-parameters, such that $\forall i$, $\lambda_i \in [0, 1]$ and $\sum_{i = 0}^{T} \lambda_{i} = 1$. With the above scheme, we observe that the resultant model remembers adversarial examples from attacks which were introduced during the initial iterations as well as recent ones, thus yielding better overall robustness.

\subsubsection{Feature Squeezing}
In addition to the above, we also investigate the suitability of Feature Squeezing as a pre-processing technique in the pipeline for improving the empirical robustness of DNNs. It is important to note that Feature Squeezing as a stand-alone defence has shown to be easily bypassed~\cite{yash2018}. When used in conjunction with our framework, in contrast, we observe that Feature Squeezing (quantization in particular) can effectively improve the accuracy against stronger attacks.
\section{Experimental Setup}
\subsection{Datasets} \label{datasets}
To measure the performance of our proposed technique, we run our experiments on three datasets standard in the computer vision/machine learning community : MNIST~\cite{mnist1998lecun}, SVHN~\cite{netzer2011svhn} and CIFAR-10~\cite{krizhevsky09learningmultiple}.

For training proxy models, we used the following datasets:
\begin{itemize}
	\item MNIST: we generated additional data using the technique described by Loosli \textit{et al.}~\cite{infinitemnist}.
    \item SVHN: we used images from the additional set of examples available in the SVHN dataset.
    \item CIFAR-10: we used images from STL-10~\cite{coates2011analysis} dataset corresponding to labels that are present in CIFAR-10. For the remaining classes (`frog`),  we picked images from Imagenet~\cite{ILSVRC15} database. All of these were down-sampled to 32$\times$32 pixels.
\end{itemize}

To ensure a fair comparison, all the target (including \textit{AMC}), as well as proxy models, were trained such that they all had accuracies within the same ballpark (Table~\ref{table:accuracies}).

\subsection{Data Preprocessing} \label{data_preprocessing}

All pixels are scaled to $[0,1]$. We employ the following split strategies for all of the three datasets:
\begin{itemize}
	\item Training data from the original dataset is used as it is for training.
    \item Test data from the original dataset is split into two portions. 30\% of this is used as validation data by the target model while training itself. Further, 30\% of the remaining 70\% data, i.e., 21\% of the original test data, is reserved while performing attacks. The remaining data is used as a test set for testing generalizability. 
    \item The attacker obtains data at its level for training itself (dataset-wise, described in Section~\ref{datasets}). That is, the data used by the proxy is independent of the target model. 
\end{itemize}

Such a data acquisition scheme closely resembles a practical scenario, where the adversary may obtain data all by itself. The data used at all stages is balanced class-wise. Data-augmentation, similar to that described by Urban \textit{et al.}, is used for CIFAR10 and SVHN to help prevent over-fitting~\cite{ubran2016extraction}.

\subsubsection{\textit{AMC} Hyper-parameters}
At the $i^{th}$ iteration, we perturb 80\% of the data using the $i^{th}$ attack, while the remaining 20\% data is perturbed equally, in a non-overlapping fashion, by all previous attacks. Using this strategy yields a ratio of 1.6625: 1.1625: 1.0375: 1 of the total perturbed data points seen by the model across all runs combined. The order of attacks used while running both variations of \textit{AMC} is: FGSM, VAP, EAP, PGM.

\subsection{Attack Hyperparameters}
Based on a trade-off between attack rates and visual inspecting perturbed images, we arrived at the hyperparameters mentioned in Table~\ref{table:attack_params} for all our experiments. All adversarial data was clipped in the range [0,1], since all the trained models expect input in this range.

\begin{table}[ht!]
\centering
\begin{tabular}{|c|c|c|c|} 
 \hline
 Attack & MNIST & SVHN & CIFAR10\\
 \hline\hline 
 \multicolumn{4}{|c|}{FGSM} \\
 \hline
 eps & 1e-1 & 1e-1 & (3e-1,6e-2) \\
 \hline
 \multicolumn{4}{|c|}{EAP} \\
 \hline
 beta & 1e-2 & 1e-2 & 1e-2\\
 binary\_steps & (5,7) & (1,3) & (1,9)\\
 max\_iterations & (8,15) & (5,10) & (5,1e3)\\
 initial\_const & 1e-3 & 3e-1 & (1e-1,1e-3)\\
 learning\_rate & 1e-1 & 2e-1 & (1e-1,1e-2)\\
 \hline
 \multicolumn{4}{|c|}{PGM} \\
 \hline
 eps & 3e-1 & 1e-1 & (1e-1,3e-2)\\
 nb iter & (15,20) & 5 & (5,40)\\
 \hline
 \multicolumn{4}{|c|}{VAP} \\
 \hline
 xi & 1 & (1e-6,1e-4) & 1e-6 \\
 num\_iters & (6,10) & (1,3) & 1 \\
 eps & (5,8) & (2,3) & 2 \\
 \hline 
\end{tabular}
\caption{Attack hyperparameters for white-box and black-box attacks used throughout in experiments. Hyperparameter names here refer to the ones in the Cleverhans library. Tuples indicate parameters for (white-box,black-box), whereas single entries signify the same hyperparameter for both cases.}
\label{table:attack_params}
\end{table}

\subsection{Evaluation Setup}
We evaluate the effectiveness of our approach against adaptive black-box adversaries as follows:
\begin{enumerate}
	\item We train two local proxy models ($\textrm{P}'$ \& $\textrm{P}''$): ($\mathrm{P}'$) is used to strengthen (or harden) the target model and ($\mathrm{P}''$) is used to measure the robustness of the hardened models to adversarial attacks. These models are replicas of each other since they have the same architecture and are trained over the same dataset.
    \item To test the effectiveness of the model hardening algorithms, we generate adversarial examples for the local proxy model $\mathrm{P}''$ using white-box strategies and attack the target model with these examples.
\end{enumerate}

\begin{table}[ht!]
\centering
\begin{tabular}{|c|c|c|c|} 
 \hline
 Model & MNIST & SVHN & CIFAR10 \\
 \hline\hline
 Undefended & 0.9 & 3.3 & 9.5\\ 
 Adv. Training & 1.4 & 2.7 & 10.4\\ 
 Adv. Training [P] & 1.1 & 3.2 & 11.7\\ 
  \hline
 \textit{AMC},target-hardened & 1.4 & 1.3 & 9.3\\
 \textit{AMC},local proxy & 1.1 & 3.5 & 9.9\\
 \hline
 Proxy & 0.2 & 6.0 & 17.1\\
 \hline
\end{tabular}
\caption{Error rates (lower is better) for various target and proxy models trained by us. Proxy here corresponds to the proxy trained with access to only class-labels returned by the target model. Adv. Training and Adv. Training[P] signify accuracies for models trained with adversarial training and proxy-based adversarial training, averaged over the four attacks.}
\label{table:accuracies}
\end{table}

\subsubsection{Proxy Model Architecture}
The architecture of the proxy model comprises of 4 convolutional layers and 2 dense layers, using \emph{ReLU} activations and dropout [0.4, 0.3, 0.2], along with $2\times2$ max pooling after every 2 convolutional layers. This architecture is similar to the one mentioned in~\cite{ubran2016extraction}. The same architecture is used for all three datasets, with changes in input shape accordingly. 

\subsubsection{Black Box Model Architectures}
For CIFAR-10 and SVHN, we used ResNet32~\cite{he2016deep} while for MNIST, we used LeNet~\cite{lecun1998gradient}.

\section{Empirical Results} \label{emperical_results}

We conducted multiple experiments, proving the efficiency of our method against several attacks, both in white-box and black-box setups. \textit{AMC, target-hardened} uses adversarial examples generated from the target model and \textit{AMC, local proxy} generates them using a local proxy. We evaluated the performance of our approach on all the three datasets listed above, for popular white-box attack algorithms such as FGSM, PGM, and compared them with both variations of our proposed framework (Section~\ref{amc}). Accuracies for the models used in our experiments are given in Table~\ref{table:accuracies}. Note that there is a slight drop in accuracies for the case of adversarial hardening, which has been reported widely in literature~\cite{madry2017pgm,dhillon2018stochastic}. Despite this fact, models trained with \textit{AMC} have generalization accuracies as good as an undefended model, which is yet another advantage over plain adversarial training.

The drop in accuracy for the proxy model on CIFAR-10 can be attributed to the significant difference in the distributions of data used by the proxy and target models. Parameters for the attacks we tested were decided upon after analyzing the adversarial images produced, using them. We visually inspected some of the generated examples to make sure that they are not just noise, while at the same time trying to maximize error induced in the target model\footnote{We used Cleverhans (https://cleverhans.readthedocs.io/en/latest/) for implementing these attacks and Tensorflow (https://www.tensorflow.org/) for training models}.

We compare our proposed framework (both variations mentioned in Section~\ref{amc}) with the best adversarially trained model in terms of empirical robustness for each of the attack.

\subsection{White-Box Attacks}
We observe that \textit{AMC, target-hardened}, on an average, gives higher empirical robustness (\textit{accuracy on adversarial examples}). Average robustness here refers to the robustness against all four attacks averaged together. We also observed that models obtained via adversarial hardening against only one kind of attack did not improve robustness against other attacks; whereas our models are robust against all the attacks we considered (Table~\ref{table:attack_numbers}).

\begin{table}[h!]
	\small
	\centering
	\begin{tabular}{ |c||c|c|c|c| }
		\hline
        \multirow{2}{*}{Model} & \multicolumn{4}{|c||}{White-Box Attacks} \\
        \cline{2-5}
        & F & E & P & V \\
		\hline  \hline
		 \multicolumn{5}{|c|}{MNIST} \\
		\hline
		Undefended & 85.4 & 86.6 & 88.9 & 69.3\\
		FGSM & \underline{5.5} & 51.2 & 65.9 & 42.4\\
		EAP & 74.2 & \underline{15.4} & 66.4 & 50.4\\
		PGM & 31.6 & 27.0 & \underline{4.4} & 33.1\\
		VAP & 27.6 & 40.6 & 64.6 & \underline{19.6}\\
		\textit{AMC},target-hardened & \textbf{5.1} & \textbf{8.4} & \textbf{3.1} & \textbf{3.2}\\
		\hline
		\multicolumn{5}{|c|}{SVHN} \\
		\hline
		Undefended & 75.7 & 95.0 & 97.6 & 34.6\\
        FGSM & \underline{2.7} & 94.2 & 94.5 & 36.9\\
        EAP & 91.6 & \underline{8.4} & 91.56 & 89.9\\
        PGM & 31.6 & 58.3 & \underline{29.3} & 49.3\\
        VAP & 26.7 & 83.6 & 94.3 & \underline{15.8}\\
		\textit{AMC},target-hardened & \textbf{2.4} & \textbf{4.9} & \textbf{25.1} & \textbf{4.5}\\
		\hline
		\multicolumn{5}{|c|}{CIFAR10} \\
		\hline
		Undefended & 86.2 & 94.4 & 97.2 & 94.4\\
		FGSM & \underline{12.0} & 94.5 & 95.1 & 36.5\\
		EAP & 90.0 & \underline{30.1} & 96.7 & 50.5\\
		PGM & 30.7 & 35.4 & \underline{63.5} & 29.3\\
		VAP & 28.9 & 70.5 & 92.1 & \underline{13.0}\\
		\textit{AMC},target-hardened & \textbf{10.6} & \textbf{23.9} & \textbf{62.5} & \textbf{11.0}\\
		\hline
	\end{tabular}
	\caption{Error rates (lower is better) for various defenses against white-box attacks. We can see \textit{AMC},target-hardened performing better than adversarial training. Attacks \{F, E, P, V\} here correspond to \{FGSM, EAP, PGM, VAP\} respectively. It is important to note that we compare our model (\textit{AMC}) with the best model in terms of empirical robustness for each of the attack.}
	\label{table:attack_numbers}
\end{table}

\subsection{Black-Box Attacks}

For the adversary's proxy models, we consider \text{adaptive} proxies trained using three possible predictive interfaces. Given an example $x$, the target models predictive interface returns:
 \begin{enumerate}
 \item only the most likely label $\hat{y}$,
 \item only the most likely label $\hat{y}$ and adds label noise to it (to approximate distillation), or
 \item a vector of tuples containing class conditional probabilities $p(y_i|x)$ and the corresponding label $y_i$.
 \end{enumerate}

Even in the case of black-box attacks, irrespective of the predictive interface made available by the target model, we observe trends similar to that of white-box attacks. The first type of proxy, i.e. the one that assumes only class-labels, is under the most generic scenario and thus best resembles a real-world black box setting. Results for this proxy are summarized in Table~\ref{table:attack_numbers_black}. We observed similar trends for the other two kinds of proxy models. Robustness of models trained with \textit{AMC} (\textit{AMC,local proxy}) is, on an average, higher than adversarial training.

\begin{table}[h!]
	\small
	\centering
	\begin{tabular}{ |c||c|c|c|c| }
    \hline
        \multirow{2}{*}{Model} & \multicolumn{4}{|c||}{Black-Box Attacks} \\
        \cline{2-5}
        & F & E & P & V \\
		\hline  \hline
		 \multicolumn{5}{|c|}{MNIST} \\
		\hline
		Undefended & 83.2 & 8.7 & 81.9 & 30.9\\
		FGSM[P] & \textbf{17.2} & 3.68 & 25.1 & 21.3\\
		EAP[P] & 82.3 & \underline{1.5} & 52.3 & 28.8\\
		PGM[P] & 52.6 & 3.78 & \underline{21.3} & \underline{4.5}\\
		VAP[P] & 31.3 & 4.0 & 39.5 & 27.03\\
		\textit{AMC}, local proxy & \underline{18.2} & \textbf{1.3} & \textbf{20.8} & \textbf{2.4}\\
		\hline
		\multicolumn{5}{|c|}{SVHN} \\
		\hline
		Undefended & 72.7 & 18.8 & 71.4 & 31.4\\
		FGSM[P] & \underline{62.2} & 15.1 & 67.4 & 27.4\\
		EAP[P] & 70.8 & 12.9 & 68.6 & 29.1\\
		PGM[P] & 64.3 & \underline{11.4} & \underline{54.0} & \underline{25.9}\\
		VAP[P] & 69.4 & 15.8 & 54.7 & 26.8\\
		\textit{AMC}, local proxy & \textbf{55.3} & \textbf{8.8} & \textbf{44.5} & \textbf{19.9}\\
		\hline
		\multicolumn{5}{|c|}{CIFAR10} \\
		\hline
	    Undefended & 66.9 & 16.0 & 44.6 & 36.4\\
		FGSM[P] & \underline{65.4} & 16.0 & 43.7 & 35.1\\
		EAP[P] & 66.6 & \textbf{16.0} & 44.2 & 35.4\\
		PGM[P] & 65.6 & 18.3 & \underline{43.6} & 35.1\\
		VAP[P] & 65.8 & 16.2 & 43.7 & \underline{34.8}\\
		\textit{AMC}, local proxy & \textbf{59.2} & \underline{16.1} & \textbf{41.3} &  \textbf{30.7}\\
		\hline
	\end{tabular}
	\caption{Error rates (lower is better) for various defenses against black-box attacks. The tag [P] signifies hardening with examples generated with a local proxy. We can see \textit{AMC}, local proxy performing better than proxy-based adversarial training. Attacks \{F, E, P, V\} here correspond to \{FGSM, EAP, PGM, VAP\} respectively. It is important to note that we compare our model (\textit{AMC}) with the best model in terms of empirical robustness for each of the attack.}
	\label{table:attack_numbers_black}
\end{table}

\subsection{Variations of \textit{AMC}}

To assert the importance of parameter transfer across iterations in our \textit{AMC} algorithm, we also ran \textit{AMC} with the configuration where there is no parameter transfer across the models in the cascade, i.e., ($\textrm{M}^{i + 1} \leftarrow \theta^{\textrm{M}^{0}}$). As expected, we observe the performance of the model to be dismal in the above case.

Since the algorithm processes attacks across cascades sequentially, one would expect this order to make a difference. To test this intuition, we tried two orders: sorted and reverse sorted by the power of attacks observed in the literature, i.e., FGSM, VAP, EAP, PGM and its reverse order. We observed higher robustness for models trained with the first order. Even though the ratio of attack data seen across runs is more-or-less equal, it is slightly higher for attacks seen earlier on in the algorithm. Scheduling stronger attacks first seems more intuitive, but it also adds the possibility of over-fitting to that attack.

\subsection{Generalizing to an unseen Attack}
\subsubsection{Feature Squeezing}
 Feature Squeezing~\cite{xu2017feature} is an input pre-processing technique that re-scales inputs to get rid of high frequency, potentially adversarial, noise. It has been shown to help achieve slightly nudged accuracies on adversarial inputs. To evaluate the compatibility and advantage of Feature Squeezing on top of models trained with \textit{AMC}, we calculate robustness numbers using this technique. (Table~\ref{table:attack_fs}). 

\begin{table}[h!]
	\small
	\centering
	\begin{tabular}{ |c||c|c|c|c| }
		\hline
        \multirow{2}{*}{Datset} & \multicolumn{2}{|c||}{Normal} & \multicolumn{2}{|c||}{Stronger} \\
        \cline{2-5}
        & EAP & PGM & EAP & PGM \\
		\hline
        \multicolumn{5}{|c|}{MNIST} \\
        \hline
        \textit{AMC} & 8.4 & 3.1 & 89.7 & 98.8\\
        \textit{AMC} and FS & 3.1 & 2.9 & 44.6 & 6.73\\
		\hline
		\multicolumn{5}{|c|}{SVHN} \\
        \hline
		\textit{AMC} & 4.9 & 25.1 & 94.2 & 98.1\\
		\textit{AMC} and FS & 2.1 & 10.1 & 39.3 & 40.5\\
		\hline
		\multicolumn{5}{|c|}{CIFAR10} \\
        \hline
	    \textit{AMC} & 23.9 & 62.5 & 93.9 & 97.6\\
	    \textit{AMC} and FS & 19.2 & 49.6 & 36.9 & 38.2\\
		\hline
	\end{tabular}
	\caption{Error rates (lower is better) for \textit{AMC} with Feature Squeezing for white-box attacks for both normal and high-attack rate hyperparameters.}
	\label{table:attack_fs}
\end{table}

We calculated these numbers for EAP and PGM attacks; for the hyper-parameters used in earlier sections (Table~\ref{table:attack_params}) as well as higher parameters (doubled 'nb\_iter' and 'eps' for PGM, 'binary\_steps' and 'max\_iterations' for EAP) to achieve higher error rates. It is important to note that Feature Squeezing as a stand-alone defence has shown to be easily bypassed~\cite{yash2018}. When used in conjunction with our framework, in contrast, we observe that Feature Squeezing (quantization in particular) can effectively improve the accuracy against stronger attacks, samples from which weren't previously seen by the model.

\subsubsection{Leave one attack out validation}
To study the capability of models trained via \textit{AMC} to generalize, we test it against an unseen attack: we use ${n-1}$ attacks for running \textit{AMC}, and the ${n^{th}}$ one for performing attacks. We experimented using FGSM as the unseen attack in a white-box setting. For all datasets, models trained with \textit{AMC} outperformed undefended models significantly in terms of robustness (Table \ref{table:attack_numbers_one_held}). These numbers are, on an average, better than all of the defence methods (apart from hardening against FGSM, since in that case the attack is known before-hand) in Table~\ref{table:attack_numbers}.

\begin{table}[h!]
	\small
	\centering
	\begin{tabular}{ |c|c| }
		\hline
        Model & FGSM White-box \\
		\hline  \hline
		 \multicolumn{2}{|c|}{MNIST} \\
		\hline
		Undefended & 85.4 \\
		\textit{AMC}(with PGM, EAP and VAP) & \underline{15.4} \\
		\textit{AMC}(with PGM, EAP and VAP) and FS & \textbf{14.8} \\
		\hline
		\multicolumn{2}{|c|}{SVHN} \\
		\hline
		Undefended & 75.7  \\
		\textit{AMC}(with PGM, EAP and VAP) & \underline{12.1}\\
		\textit{AMC}(with PGM, EAP and VAP) and FS & \textbf{11.0} \\
		\hline
		\multicolumn{2}{|c|}{CIFAR10} \\
		\hline
	    Undefended & 86.2 \\
		\textit{AMC}(with PGM, EAP and VAP) & \underline{24.7} \\
		\textit{AMC}(with PGM, EAP and VAP) and FS & \textbf{23.9} \\
		\hline
	\end{tabular}
	\caption{Error rates (lower is better) for models trained with \textit{AMC} on PGM, VAP and EAP in a white-box setting. When compared against the undefended model, we can see that \textit{AMC} is performing better against FGSM attack, which it hasn't seen during training. Using Feature Squeezing (FS) in conjunction with AMC further brings the error rates down.}
	\label{table:attack_numbers_one_held}
\end{table}
\section{Discussion}

As we observed in our experiments, there is no defence technique that works against all adversarial attacks which are effective for complex datasets such as SVHN and CIFAR10; hardening a model for a specific attack does not necessarily lead to an increase in robustness against future attacks of that kind. Adversarial training, as we observed, increased robustness only against the attack it is hardened against and did not substantially increase robustness against other attacks significantly.

The key contributions of the \textit{AMC} framework proposed in the paper are as follows :
\begin{enumerate}
	\item \textit{AMC} provides robustness against all the attacks it is hardened against, making it an all-in-one defence mechanism against several attacks. 
	\begin{itemize}
		\item In the case of white-box attacks, on an average \textit{AMC} provides an absolute increase in robustness of $6.225\%$ for MNIST, $5.075\%$ for SVHN and $2.65\%$ for CIFAR10, in comparison to adversarial hardening.
		\item In the case of black-box attacks, on an average \textit{AMC} provides an absolute increase in robustness of $0.45\%$ for MNIST, $6.25\%$ for SVHN and $2.7\%$ for CIFAR10, in comparison to local-proxy based adversarial hardening.
	\end{itemize}
	\item Even though Feature Squeezing has been shown to be bypassed, when used in conjunction with our framework, we observe that Feature Squeezing (quantization in particular) can effectively improve the robustness against both attacks seen by the model during the training (on an average, an absolute increase of $2.65\%$ for MNIST, $8.9\%$ for SVHN and $8.8\%$ for CIFAR10) and stronger attacks which weren't previously seen by the model (on an average, an absolute increase of $68.9\%$ for MNIST, $56.25\%$ for SVHN and $58.2\%$ for CIFAR10) on top of \textit{AMC}.
	\item There is no overhead at inference time. Thus, services can deploy versions of the models hardened with \textit{AMC} without any compromise on latency.
	\item It is easy to incorporate into already trained models. As the intermediate step in building the model cascades involves fine-tuning , it is much faster than existing defensive methods which require training from scratch for every attack.
	\item As opposed to the performance of models from Madry \textit{et. al.}~\cite{madry2017pgm} the resultant model does not compromise on predictive performance on the unseen test set (Table~\ref{table:accuracies}).
	\item We also observe that the performance of AMC for unseen attacks is comparable to models hardened against those specific attacks (Table~\ref{table:attack_numbers_one_held}). An interesting direction for future would be to evaluate the performance of \textit{AMC} over a larger set of unseen attacks.
\end{enumerate}

To the best of our knowledge, ours is the first attempt to provide an end to end pipeline to improve robustness against multiple adversarial attacks simultaneously; in both white-box and black-box settings. 


\clearpage
\bibliography{references}
\bibliographystyle{aaai}
\end{document}